\documentclass[letterpaper, 10 pt, conference]{ieeeconf}  

\IEEEoverridecommandlockouts                              

\usepackage{amsmath,amsfonts}
\usepackage{algorithmic}
\usepackage{array}
\usepackage[caption=false,font=normalsize,labelfont=sf,textfont=sf]{subfig}
\usepackage{textcomp}
\usepackage{stfloats}
\usepackage{url}
\usepackage{verbatim}
\usepackage{graphicx}
\usepackage{booktabs}
\usepackage{pgfplots}
\usepackage{pgf-pie}
\usepackage{multirow}

\usepackage{booktabs}
\usepackage{amssymb}

\usepackage{graphics} 
\usepackage{epsfig} 
\usepackage{mathptmx} 
\usepackage{times} 
\usepackage{amsmath} 
\usepackage{amssymb}  

\usepackage{booktabs}   
\usepackage{multirow}   
\usepackage{array}      
\newcolumntype{M}[1]{>{\raggedright\arraybackslash}m{#1}}

\newcolumntype{C}[1]{>{\centering\arraybackslash}m{#1}}

\def\BibTeX{{\rm B\kern-.05em{\sc i\kern-.025em b}\kern-.08em
    T\kern-.1667em\lower.7ex\hbox{E}\kern-.125emX}}

\usepackage{balance}

\usepackage{booktabs}
\usepackage{tabularx}
\usepackage{array}

\newcolumntype{Y}{>{\centering\arraybackslash}X}

\makeatletter
\let\NAT@parse\undefined
\makeatother

\usepackage[hidelinks]{hyperref}

\title{\LARGE \bf
InterScenes: A Multimodal Interaction-Evolution Dataset for VLA Evaluation and Safety Validation in Automated Driving
}

\author{
Mengjie Tian$^{1}$,
Xinrui Zhang$^{1}$,
Peizhi Zhang$^{1,*}$,
Guirong Zhuo$^{1}$,
Zhuoren Li$^{2}$,\\
Qixiang Zhang$^{3}$,
Junpeng Huang$^{1}$,
Haojie Feng$^{1}$,
and Lu Xiong$^{1}$%
\thanks{$^{1}$School of Automotive and Energy Engineering,
Tongji University, China.
$^{2}$Department of Civil Engineering,
The University of Hong Kong, China.
$^{3}$School of Mechanical Engineering,
Southeast University, China.
$^{*}$Corresponding author: Peizhi Zhang
(zhangpeizhitom@126.com).}
}

\begin{document}

\maketitle
\thispagestyle{empty}
\pagestyle{empty}

\begin{abstract}
Understanding and responding to dynamic interactions is fundamental to automated driving systems (ADSs), particularly end-to-end and vision-language-action (VLA) models. However, many existing interaction-oriented datasets primarily support
short-horizon motion forecasting and rarely characterize the temporal
evolution of traffic interactions, including conflict formation, risk
escalation, and eventual resolution. Moreover, these datasets are often trajectory-centric
and lack synchronized onboard multimodal observations and physically
grounded descriptions of interaction evolution, limiting their utility
for VLA evaluation and safety validation. To address these limitations, we introduce InterScenes, a real-world multimodal interaction-evolution dataset for VLA evaluation and safety validation. InterScenes represents traffic interactions as temporal processes and provides synchronized multi-view images, LiDAR point clouds, ego and surrounding-agent trajectories, and physically grounded interaction-evolution descriptions. We develop a trajectory-enhanced vision-language model annotation pipeline integrating visual observations with trajectory-derived physical cues to automatically generate these descriptions. The dataset contains 1,190 interaction episodes spanning
diverse road environments, traffic participants, and
interaction types, including interactive takeovers,
with 24.2\% of episodes exhibiting a minimum
time-to-conflict-point difference below 3~s. We further demonstrate the utility of InterScenes through VLA-based planning evaluation and semantic-guided scenario generation. Cross-dataset evaluation shows that InterScenes poses greater challenges to VLA-based trajectory planning, while physically grounded interaction-evolution descriptions improve the physical fidelity of semantic-guided scenario generation
in simulation and support proving-ground deployment for ADS safety validation. Project page:
\url{https://autonomous-dataset.github.io/InterScenes-project-page/}.
\end{abstract}

\section{Introduction}

Understanding and responding to evolving traffic
interactions is a fundamental capability of advanced automated driving
systems (ADSs), particularly end-to-end (E2E) and vision-language-action
(VLA) models~\cite{liu2025autonomous,ARTEMIS-3,zhaoyongqisurvey,XU2026116002}.
Driving decisions in complex traffic environments are shaped by the temporal
evolution of behaviors and interactions among surrounding participants,
rather than by instantaneous states alone. Effective decision-making therefore depends on understanding surrounding agents' evolving behaviors and interaction intentions, while continuously adapting to unfolding interactions~\cite{Drivellm,Driveanywhere}.
Consequently, evaluating ADS understanding and response to evolving interactions has become a critical challenge for evaluation and safety validation.

Existing driving datasets have supported substantial progress in perception,
prediction, planning, and interaction analysis. Multimodal datasets such as
nuScenes~\cite{nuScenes} and the Waymo Open Dataset~\cite{WOD} provide
rich onboard sensor observations, while trajectory- and interaction-oriented
datasets such as INTERACTION~\cite{INTERACTION}, WOMD~\cite{WOMD},
SinD~\cite{SIND}, and InterHub~\cite{interhub} provide detailed
multi-agent motion. More recent E2E- and VLA-oriented datasets further
extend these capabilities toward driving planning and decision-making
~\cite{wod_e2e,VLA4CoDrive}. However, these datasets are generally
organized around perception, motion forecasting, interaction prediction,
or driving decisions rather than explicitly representing interaction
evolution as a complete process. Even when temporally extended sequences
are available, the semantic and physical evolution of an interaction is
rarely explicitly characterized throughout its development and resolution.
This makes it difficult to evaluate ADS behavior throughout the
interaction process.

A second limitation concerns the information available for interaction
understanding and evaluation. Trajectory-centric datasets~\cite{INTERACTION,WOMD,SIND,interhub} provide detailed multi-agent
motion but often lack synchronized onboard observations and visual context.
In contrast, multimodal, E2E-, and VLA-oriented datasets
~\cite{nuScenes,WOD,wod_e2e,VLA4CoDrive} provide richer sensor
observations and planning supervision, but generally lack physically
grounded descriptions of interaction evolution. Non-standard scenario
sources, such as CCD~\cite{ccd}, NHTSA SGO Reports~\cite{nhtsa_sgo},
and TrafficAlign~\cite{trafficalign}, broaden the coverage of diverse
and safety-relevant scenarios but typically lack a unified multimodal and
trajectory-based representation. As summarized in
Table~\ref{tab:dataset_comparison}, multimodal observations, multi-agent
trajectories, interaction-evolution annotations, and takeover events are
rarely available together within the same data source. This makes it difficult to jointly support both VLA evaluation and
scenario-based ADS safety validation through a unified data representation.

\begin{table*}[t]
\centering
\vspace*{3mm}
\caption{Comparison of representative driving datasets and scenario sources.}
\label{tab:dataset_comparison}

\scriptsize
\renewcommand{\arraystretch}{1.12}
\setlength{\tabcolsep}{2.4pt}

\begin{tabularx}{\textwidth}{
@{}
>{\raggedright\arraybackslash}m{2.55cm}
>{\centering\arraybackslash}m{0.55cm}
>{\raggedright\arraybackslash}m{1.40cm}
>{\raggedright\arraybackslash}m{1.40cm}
YYYYYYY
@{}}
\toprule

Dataset / Source &
Year &
Origin &
Viewpoint &
\shortstack{Interaction-\\Centered} &
\shortstack{Multi-Agent\\Trajectories} &
\shortstack{RGB\\Images} &
LiDAR &
\shortstack{Interaction-\\Evolution\\Annotation} &
\shortstack{VLA Planning\\Support} &
\shortstack{Takeover\\Events} \\

\midrule

CCD~\cite{ccd}
& 2020
& Derived
& Onboard
& P & -- & $\checkmark$ & -- & -- & -- & -- \\

NHTSA SGO Reports~\cite{nhtsa_sgo}
& 2022
& Report-based
& N/A
& P & -- & -- & -- & P & -- & P \\

TrafficAlign~\cite{trafficalign}
& 2026
& Derived
& Onboard
& -- & -- & $\checkmark$ & -- & -- & -- & -- \\

\addlinespace[1pt]
\cmidrule{1-11}
\addlinespace[1pt]

INTERACTION~\cite{INTERACTION}
& 2019
& Self-collected
& Aerial/Roadside
& $\checkmark$ & $\checkmark$ & -- & -- & -- & -- & -- \\

WOMD~\cite{WOMD}
& 2021
& Self-collected
& Onboard
& $\checkmark$ & $\checkmark$ & -- & -- & -- & -- & -- \\

SinD v1.0~\cite{SIND}
& 2022
& Self-collected
& Aerial
& -- & $\checkmark$ & -- & -- & -- & -- & -- \\

InterHub~\cite{interhub}
& 2024
& Derived
& N/A
& $\checkmark$ & $\checkmark$ & -- & -- & -- & -- & -- \\

\addlinespace[1pt]
\cmidrule{1-11}
\addlinespace[1pt]

nuScenes~\cite{nuScenes}
& 2019
& Self-collected
& Onboard
& -- & $\checkmark$ & $\checkmark$ & $\checkmark$
& -- & $\checkmark$ & -- \\

Waymo Open Dataset~\cite{WOD}
& 2019
& Self-collected
& Onboard
& -- & $\checkmark$ & $\checkmark$ & $\checkmark$
& -- & $\checkmark$ & -- \\

WOD-E2E~\cite{wod_e2e}
& 2025
& Self-collected
& Onboard
& P & -- & $\checkmark$ & --
& -- & $\checkmark$ & -- \\

VLA4CoDrive~\cite{VLA4CoDrive}
& 2026
& Synthetic
& Onboard
& P & $\checkmark$ & $\checkmark$ & $\checkmark$
& -- & $\checkmark$ & -- \\

\addlinespace[1pt]
\midrule

\textbf{InterScenes}
& 2026
& Self-collected
& Onboard
& $\checkmark$
& $\checkmark$
& $\checkmark$
& $\checkmark$
& $\checkmark$
& $\checkmark$
& $\checkmark$ \\

\bottomrule
\end{tabularx}

\vspace{1mm}
\begin{minipage}{0.99\textwidth}
\scriptsize
\textit{Notes:}
Year denotes the first public release.
$\checkmark$ denotes explicitly provided, P denotes partially available
or subset-specific, and -- denotes not explicitly provided.
Only released modalities are counted.
Interaction-Evolution Annotation requires temporally ordered semantic
characterization of interaction development.
VLA Planning Support requires synchronized onboard RGB observations
and metric future ego-motion or action ground truth.
\end{minipage}

\end{table*}

To address these gaps, we introduce \textbf{InterScenes}, a real-world
multimodal interaction-evolution dataset that jointly provides synchronized
sensor observations, multi-agent trajectories, and physically grounded
descriptions of how traffic interactions evolve over time. By representing
each interaction as a continuous temporal process, InterScenes supports
both VLA-based planning evaluation and ADS safety validation. Its physically
grounded interaction descriptions further provide constraints for
semantic-guided scenario generation and proving-ground deployment.
The main contributions are summarized as follows:

\begin{itemize}
        \item We introduce InterScenes, a real-world multimodal
    interaction-evolution dataset covering diverse road topologies,
    traffic participants, interaction types, and interactive takeovers.
    Of these episodes, 11.6\% and 24.2\% exhibit a minimum
    time-to-conflict-point difference below 1~s and 3~s,
    respectively, suggesting substantial coverage of
    potentially high-risk interaction scenarios.

    \item We develop a trajectory-enhanced vision-language model (VLM) annotation pipeline
    that integrates visual observations with trajectory-derived physical
    cues to generate physically grounded interaction-evolution descriptions.

    \item We demonstrate the utility of InterScenes for VLA-based planning
    evaluation and ADS safety validation. Its interaction-evolution
    descriptions improve the physical fidelity of semantic-guided scenario
    generation and support proving-ground deployment.
\end{itemize}

\section{InterScenes Construction}

\begin{figure*}[!t]
\centering
\vspace*{1mm}
\includegraphics[width=\textwidth]{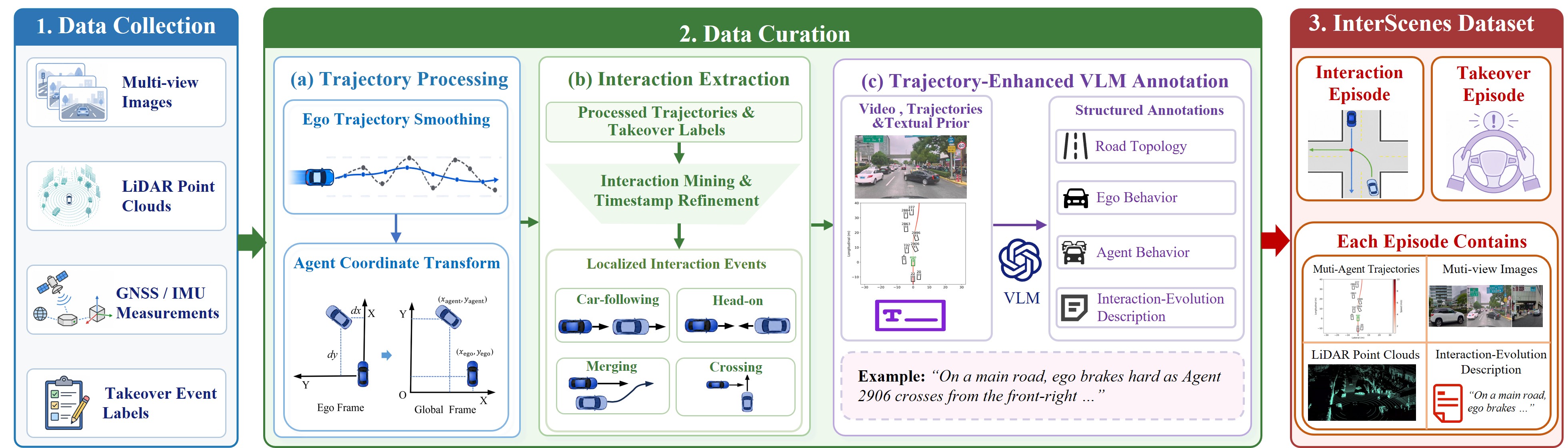}
\caption{Overview of the InterScenes construction framework, including
multimodal data collection, trajectory processing, interaction extraction,
and trajectory-enhanced VLM annotation.}
\label{fig:data_pipeline}
\end{figure*}

The construction framework of InterScenes is illustrated in
Fig.~\ref{fig:data_pipeline}, covering multimodal data collection,
trajectory processing, interaction extraction, trajectory-enhanced VLM
annotation, and dataset organization. 
Synchronized multimodal driving data are processed into
interaction-centered episodes with aligned sensor observations,
multi-agent trajectories, and interaction-evolution descriptions.

\subsection{Data Collection}
\begin{figure}[!t]
\centering
\includegraphics[width=0.85\columnwidth]{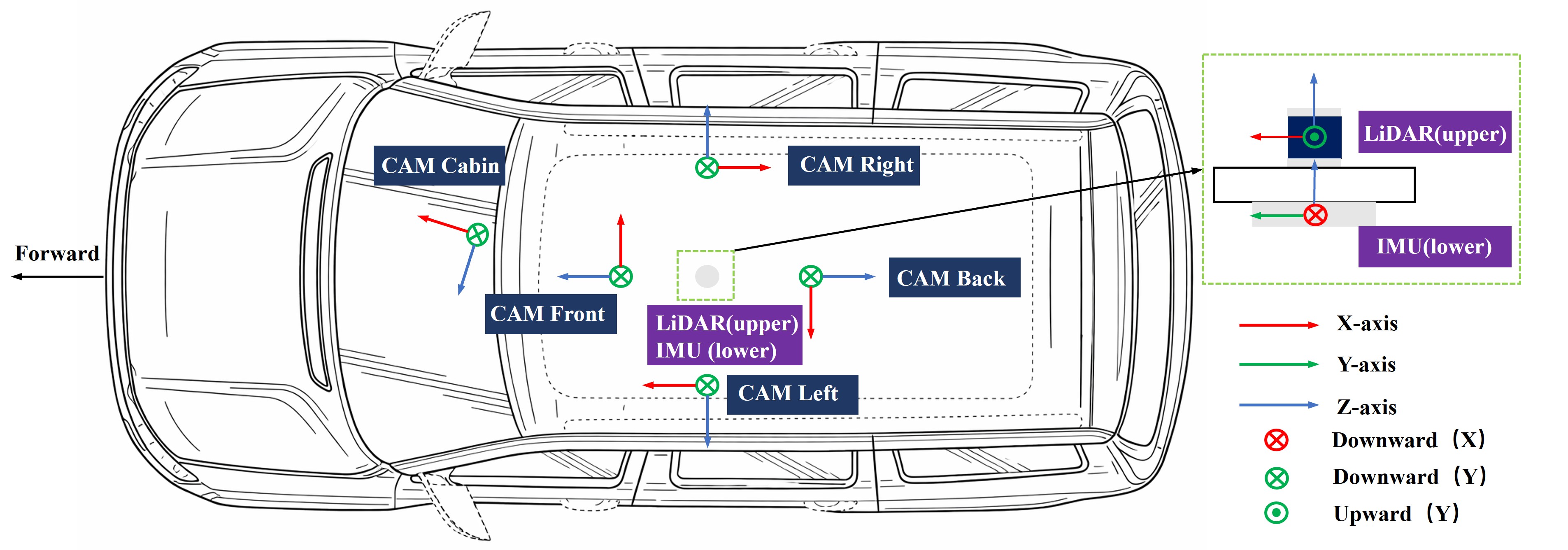}
\caption{Sensor configuration of the InterScenes data collection platform.}
\label{fig:sensor_setup}
\end{figure}

As shown in Fig.~\ref{fig:sensor_setup}, InterScenes was collected
using a production vehicle operating with its assisted-driving
function engaged. A synchronized multimodal acquisition platform
recorded images from four onboard cameras, in-cabin video,
point clouds from a RoboSense Ruby Plus 128-beam LiDAR,
and navigation measurements from a CHCNAV CGI-430 integrated
GNSS/INS system. The manufacturer-specified accuracies are
$\pm 2$~cm for LiDAR ranging and
$1~\mathrm{cm}+1~\mathrm{ppm}$ (RMS) for horizontal positioning
in RTK mode. All sensors were synchronized using a dedicated hardware
synchronization unit with pulse-per-second (PPS) timing,
achieving a synchronization error below 1~ms. The onboard cameras
covered the front, rear, left, and right views, while the
in-cabin camera captured driver actions and the dashboard.

Data were collected in Shanghai, China, primarily during morning
and evening peak hours to capture dense traffic interactions.
Routes included local roads, commercial areas, signalized
intersections, and urban expressways, covering car-following,
merging, crossing, and head-on interactions, as well as
interactions with vulnerable road users (VRUs).

\subsection{Data Curation}

Dataset curation consists of trajectory processing, interaction extraction,
and trajectory-enhanced VLM annotation.

\subsubsection{Trajectory Processing}

Ego trajectories are converted from geographic to Universal
Transverse Mercator (UTM) coordinates, cleaned for outliers,
and smoothed using a Kalman filter. Surrounding-agent positions
in the LiDAR coordinate frame and cross-frame track IDs are
obtained using a manufacturer-provided processing tool.
The agent positions are then transformed into the same global
coordinate frame using the calibrated sensor extrinsics and
ego poses, yielding unified ego and surrounding-agent trajectories.

\subsubsection{Interaction Extraction}

Ordinary interaction episodes and interactive takeover episodes
are extracted through separate procedures. Ordinary interactions
are identified from the processed global trajectories according
to the definitions of car-following, merging, crossing, and
head-on interactions in InterHub~\cite{interhub}.
For each interaction, a type-specific conflict time is computed
from the global trajectories and used as the event timestamp.
Interactive takeover events are identified from recorded
takeover labels, with their timestamps manually refined.
Both event types are segmented into 20~s episodes spanning
8~s before and 12~s after the corresponding event timestamp.

\subsubsection{Trajectory-Enhanced VLM Annotation}

\begin{figure}[!t]
    \centering
    \includegraphics[width=0.85\columnwidth]{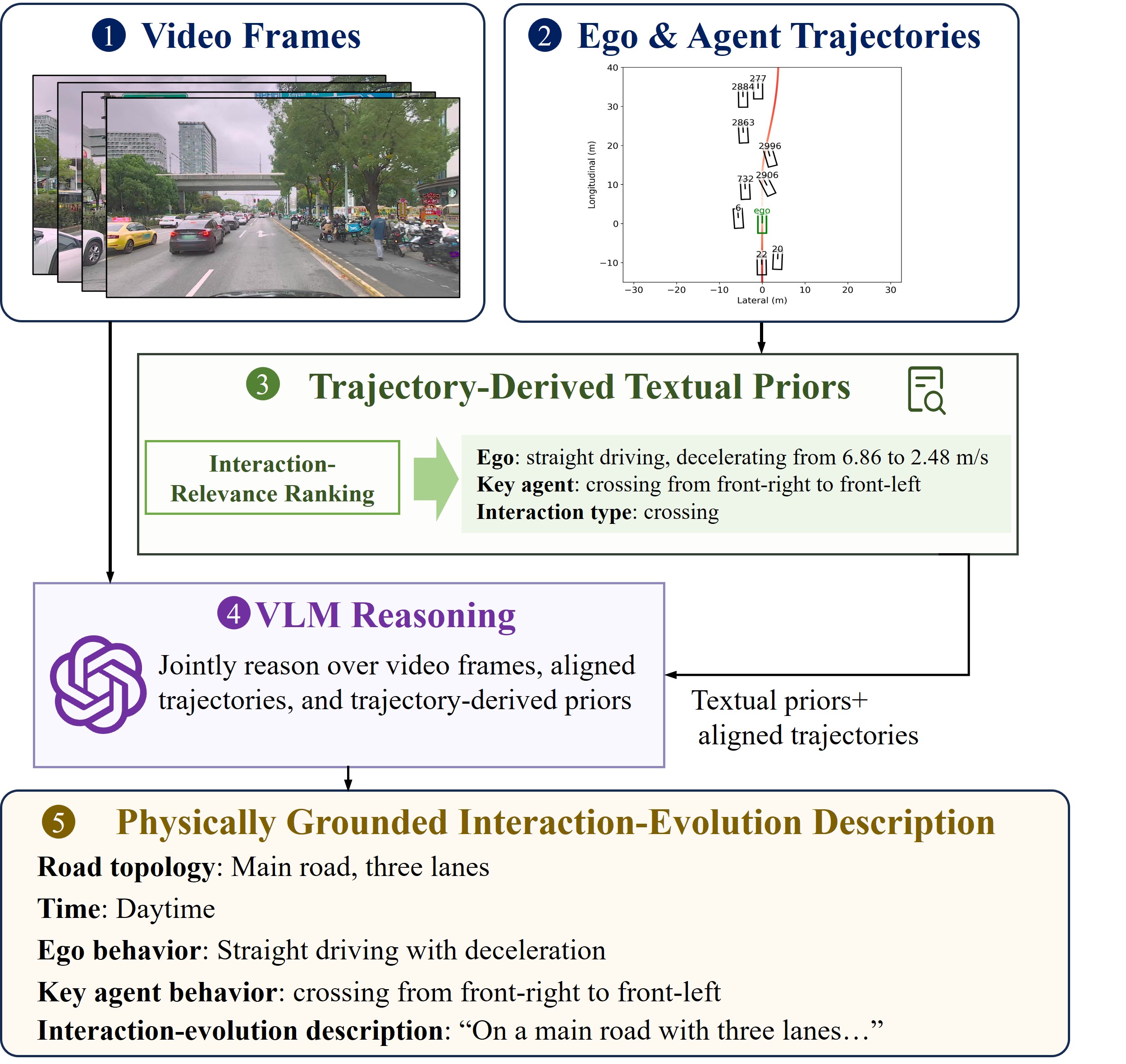}
\caption{Trajectory-enhanced VLM annotation pipeline.}
    \label{fig:vlm_annotation}
\end{figure}

As illustrated in Fig.~\ref{fig:vlm_annotation}, ego and
surrounding-agent trajectories are aligned with
visual observations for each episode. Interaction-relevant
agents are identified through trajectory-based candidate
selection and hierarchical ranking. Candidates are prioritized
using the interaction category identified during
extraction and type-specific urgency measures, with heuristic
interaction scores providing secondary ranking criteria.
Up to 10 agents are selected for textual-prior construction,
with the top-ranked agent designated as the primary
interacting agent. The textual priors summarize ego motion,
key-agent behaviors, and changes in relative position and speed.

Video frames sampled at 2~Hz, aligned trajectories of the ego
vehicle and surrounding agents, initial interaction labels,
and textual priors are incorporated into a task-specific prompt.
We use the GPT-5.4 API with
chain-of-thought (CoT)
prompting~\cite{yu2026ssquaredvladecouplingsemanticspatial,
gao2026nuriskvisualquestionanswering}
to integrate visual context with trajectory-derived physical
information. Initial interaction labels serve as guidance
and may be revised by the VLM based on evidence.
The resulting structured annotations describe road topology,
traffic-participant types, ego and surrounding-agent behaviors,
interaction type, and a physically grounded account of
interaction evolution.

\subsection{Dataset Organization}

InterScenes organizes interaction and takeover episodes using a structured
JSON-based schema that links ego and surrounding-agent states with synchronized
multi-view images, LiDAR point clouds, and semantic annotations. Each annotation
specifies road topology, lighting conditions, key interacting agents and their
behaviors, ego behavior, and physically grounded descriptions of interaction
evolution, supporting attribute-based and text-based retrieval.

\subsection{Privacy Protection and Anonymization}

All data were anonymized in accordance with relevant data-protection
requirements in China~\cite{dsl2021}. Human faces and vehicle license
plates were automatically detected and blurred~\cite{dashcam_anonymizer}.
Trajectory coordinates were further transformed to remove direct
georeferencing. Only anonymized data are included in the released dataset.

\section{Dataset Analysis}

\subsection{Dataset Diversity}
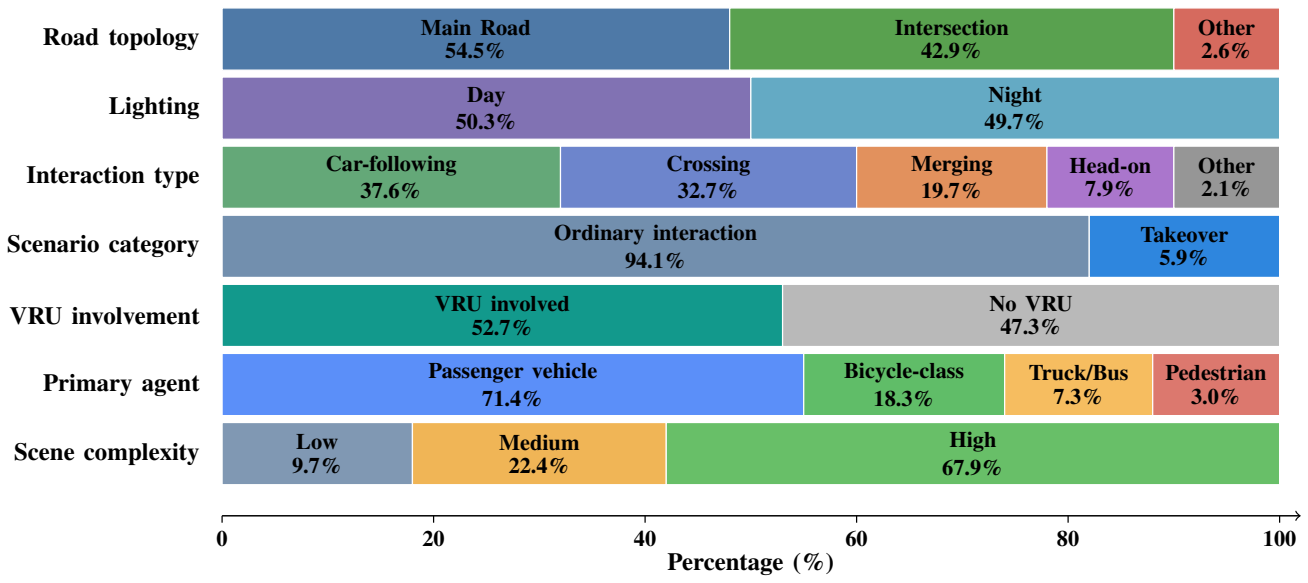
\begin{figure}[t]
\centering
\vspace{2mm}
\resizebox{\columnwidth}{!}{%
\begin{tikzpicture}[
    x=0.15cm,
    y=0.98cm,
    lab/.style={
        anchor=east,
        font=\bfseries\normalsize,
        align=right
    },
    txt/.style={
        font=\bfseries\small,
        align=center,
        text=black
    },
    txtsmall/.style={
        font=\bfseries\scriptsize,
        align=center,
        text=black
    }
]


\def\H{0.9}   

\newcommand{\BarLabel}[2]{%
    \node[lab] at (-1.3,#1+0.45) {#2};
}

\newcommand{\Seg}[5]{%
    \filldraw[
        fill=#4,
        draw=white,
        line width=0.5pt
    ]
    (#2,#1) rectangle ++(#3,\H);

    \node[txt]
    at (#2+0.5*#3,#1+0.5*\H)
    {#5};
}

\newcommand{\SegSmall}[5]{%
    \filldraw[
        fill=#4,
        draw=white,
        line width=0.5pt
    ]
    (#2,#1) rectangle ++(#3,\H);

    \node[txt]
    at (#2+0.5*#3,#1+0.5*\H)
    {#5};
}


\definecolor{cRoadMain}{RGB}{78,121,167}
\definecolor{cRoadInt}{RGB}{89,161,79}
\definecolor{cRoadOther}{RGB}{214,106,94}

\definecolor{cDay}{RGB}{129,114,179}
\definecolor{cNight}{RGB}{100,170,196}

\definecolor{cCF}{RGB}{100,168,120}
\definecolor{cCross}{RGB}{109,133,204}
\definecolor{cMerge}{RGB}{226,145,93}
\definecolor{cHead}{RGB}{170,118,204}
\definecolor{cOther}{RGB}{150,150,150}

\definecolor{cOrd}{RGB}{114,143,174}
\definecolor{cTake}{RGB}{46,134,222}

\definecolor{cVRU}{RGB}{18,153,140}
\definecolor{cNoVRU}{RGB}{185,185,185}

\definecolor{cPA1}{RGB}{91,143,249}
\definecolor{cPA2}{RGB}{96,189,104}
\definecolor{cPA3}{RGB}{246,189,96}
\definecolor{cPA4}{RGB}{220,120,110}

\definecolor{cLow}{RGB}{128,152,179}
\definecolor{cMed}{RGB}{240,181,86}
\definecolor{cHigh}{RGB}{104,191,104}


\def\yA{6.0}
\def\yB{5.0}
\def\yC{4.0}
\def\yD{3.0}
\def\yE{2.0}
\def\yF{1.0}
\def\yG{0.0}

%
%

\BarLabel{\yA}{Road topology}

\Seg{\yA}{0}{48}{cRoadMain}
    {\shortstack{Main Road\\54.5\%}}

\Seg{\yA}{48}{42}{cRoadInt}
    {\shortstack{Intersection\\42.9\%}}

\SegSmall{\yA}{90}{10}{cRoadOther}
    {\shortstack{Other\\2.6\%}}

%
%

\BarLabel{\yB}{Lighting}

\Seg{\yB}{0}{50}{cDay}
    {\shortstack{Day\\50.3\%}}

\Seg{\yB}{50}{50}{cNight}
    {\shortstack{Night\\49.7\%}}

%
%

\BarLabel{\yC}{Interaction type}

\Seg{\yC}{0}{32}{cCF}
    {\shortstack{Car-following\\37.6\%}}

\Seg{\yC}{32}{28}{cCross}
    {\shortstack{Crossing\\32.7\%}}

\Seg{\yC}{60}{18}{cMerge}
    {\shortstack{Merging\\19.7\%}}

\SegSmall{\yC}{78}{12}{cHead}
    {\shortstack{Head-on\\7.9\%}}

\SegSmall{\yC}{90}{10}{cOther}
    {\shortstack{Other\\2.1\%}}

%
%

\BarLabel{\yD}{Scenario category}

\Seg{\yD}{0}{82}{cOrd}
    {\shortstack{Ordinary interaction\\94.1\%}}

\SegSmall{\yD}{82}{18}{cTake}
    {\shortstack{Takeover\\5.9\%}}

%
%

\BarLabel{\yE}{VRU involvement}

\Seg{\yE}{0}{53}{cVRU}
    {\shortstack{VRU involved\\52.7\%}}

\Seg{\yE}{53}{47}{cNoVRU}
    {\shortstack{No VRU\\47.3\%}}

%
%
%

\BarLabel{\yF}{Primary agent}

\Seg{\yF}{0}{55}{cPA1}
    {\shortstack{Passenger vehicle\\71.4\%}}

\Seg{\yF}{55}{19}{cPA2}
    {\shortstack{Bicycle-class\\18.3\%}}

\SegSmall{\yF}{74}{14}{cPA3}
    {\shortstack{Truck/Bus\\7.3\%}}

\SegSmall{\yF}{88}{12}{cPA4}
    {\shortstack{Pedestrian\\3.0\%}}

%
%
%
%

\BarLabel{\yG}{Scene complexity}

\SegSmall{\yG}{0}{18}{cLow}
    {\shortstack{Low\\9.7\%}}

\Seg{\yG}{18}{24}{cMed}
    {\shortstack{Medium\\22.4\%}}

\Seg{\yG}{42}{58}{cHigh}
    {\shortstack{High\\67.9\%}}


\draw[->, line width=0.6pt]
    (0,-0.45) -- (102,-0.45);

\foreach \x in {0,20,40,60,80,100}{
    \draw[line width=0.5pt]
        (\x,-0.45) -- (\x,-0.55);

    \node[
        font=\bfseries\small
    ]
    at (\x,-0.78) {\x};
}

\node[
    font=\bfseries\normalsize
]
at (50,-1.15)
{Percentage (\%)};

\end{tikzpicture}%
}

\caption{Distribution of key attributes in InterScenes.}

\label{fig:interscenes_attribute_distribution}
\end{figure}

As shown in Fig.~\ref{fig:interscenes_attribute_distribution}, InterScenes
covers diverse road environments, lighting conditions, interaction types,
and traffic participants. Main-road and intersection scenarios account for
54.5\% and 42.9\% of the cases, respectively, while daytime and nighttime
conditions are nearly balanced. Car-following accounts for 37.6\% of the
interactions, crossing for 32.7\%, merging for 19.7\%, and head-on interactions for 7.9\%. Moreover, 52.7\% of the cases involve at
least one VRU, while takeover events account for 5.9\%, representing a
relatively rare subset of safety-relevant interactions. For the primary
interacting agent, defined as the top-ranked agent by the interaction-relevance
ranking, passenger vehicles account for 71.4\%, bicycle-class participants
for 18.3\%, trucks and buses for 7.3\%, and pedestrians for 3.0\%. Local
traffic complexity is defined by the number of valid non-Ego agents within
20~m of the ego vehicle at the interaction timestamp, with 0--2, 3--5, and
at least 6 agents corresponding to low, medium, and high complexity,
respectively. Under this definition, 67.9\% of the cases are categorized as
high-complexity, indicating substantial coverage of dense multi-agent
interactions.

\begin{figure}[t]
    \centering
    \includegraphics[width=\columnwidth]
    {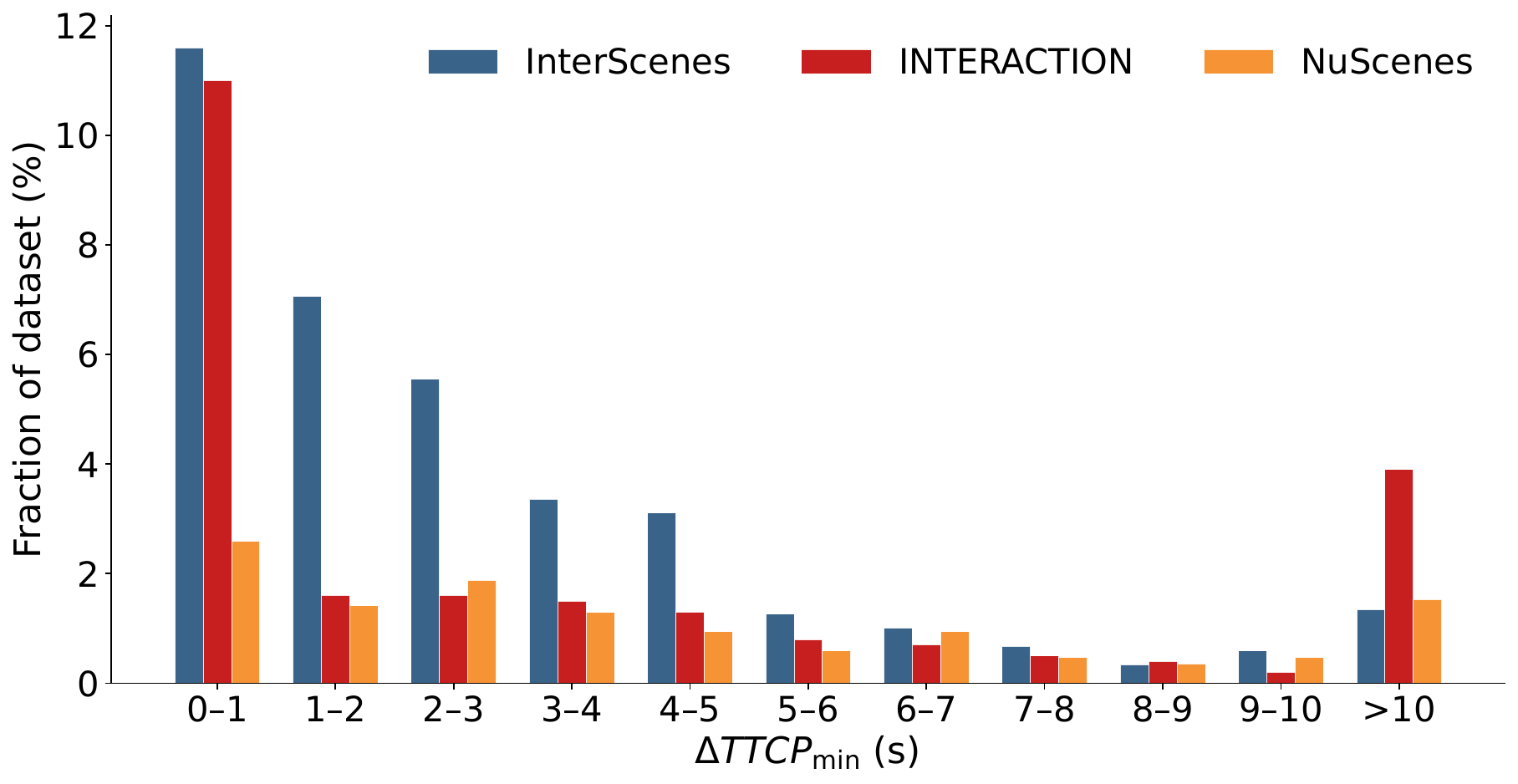}
  \caption{Comparison of conflict-point temporal proximity
across InterScenes, INTERACTION, and nuScenes using
$\Delta TTCP_{\min}$.}
    \label{fig:temporal_metric_dataset_comparison_0_10plus}
\end{figure}

\subsection{Interaction Criticality Across Datasets}
\label{subsec:interaction_criticality}


To maintain consistency with the cross-dataset interaction
analysis in INTERACTION~\cite{INTERACTION}, we adopt
the minimum time-to-conflict-point difference,
$\Delta TTCP_{\min}$, to compare interaction temporal
proximity across InterScenes, INTERACTION, and
nuScenes~\cite{nuScenes}. This metric characterizes
the difference in arrival times of interacting agents
at shared conflict points. As shown in
Fig.~\ref{fig:temporal_metric_dataset_comparison_0_10plus},
11.60\% of InterScenes cases have a $\Delta TTCP_{\min}$
below 1~s and 24.20\% below 3~s, compared with 11.00\%
and 14.20\% for INTERACTION, and 2.59\% and 5.88\%
for nuScenes, respectively. Although InterScenes and
INTERACTION show similar proportions below 1~s,
InterScenes exhibits a substantially higher proportion
in the 1--3~s range, indicating greater coverage of
temporally close interactions at shared conflict points.

\begin{figure}[t]
    \centering
    \vspace*{1mm}\includegraphics[width=0.85\columnwidth]
    {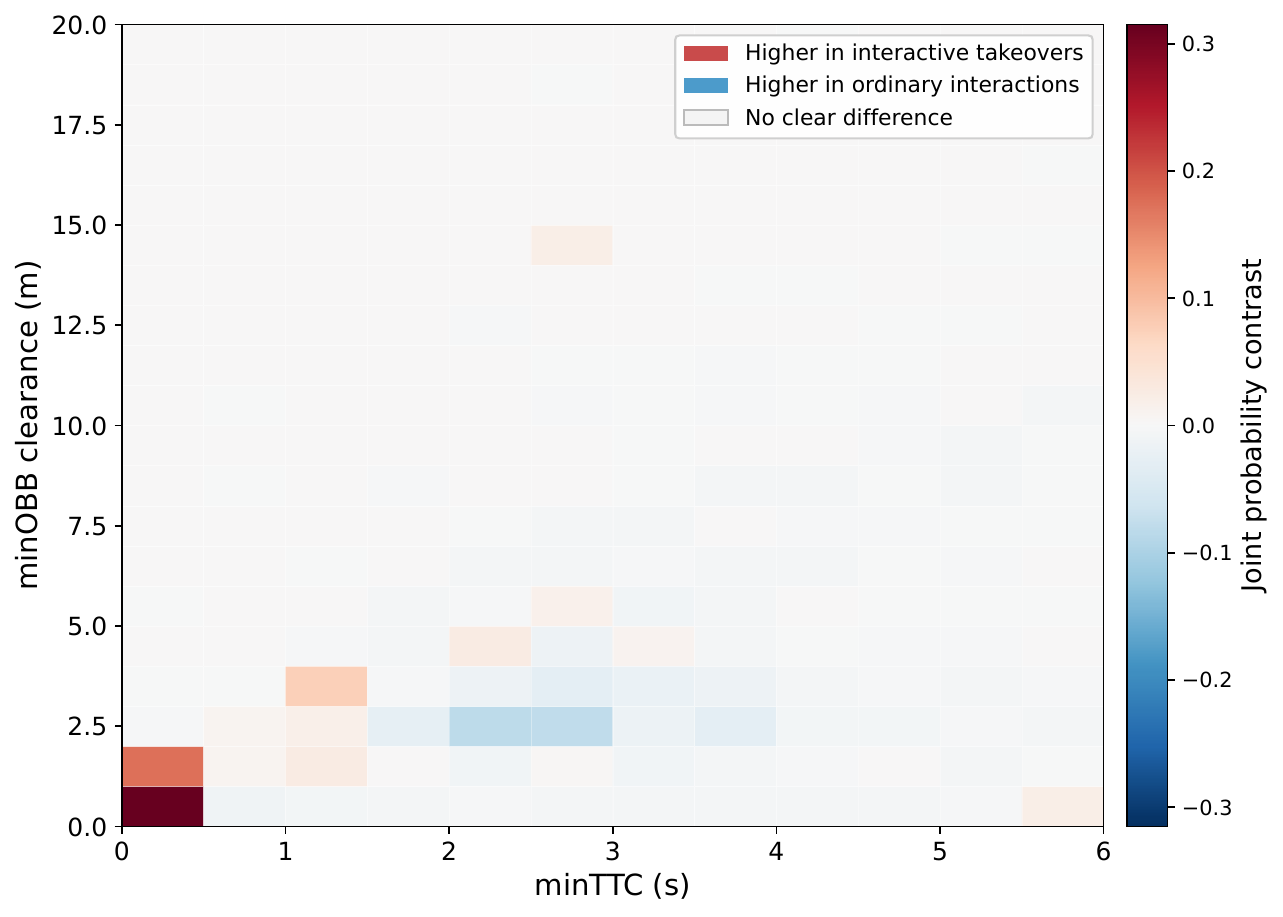}
    \caption{Difference in the joint distribution of minTTC and
    minOBB clearance between interactive takeovers and ordinary
    interactions in InterScenes.}
    \label{fig:ordinary_vs_takeover_ttc_obb}
\end{figure}

\begin{figure*}[t]
    \centering
    \vspace*{1mm}
    \includegraphics[width=\textwidth]
    {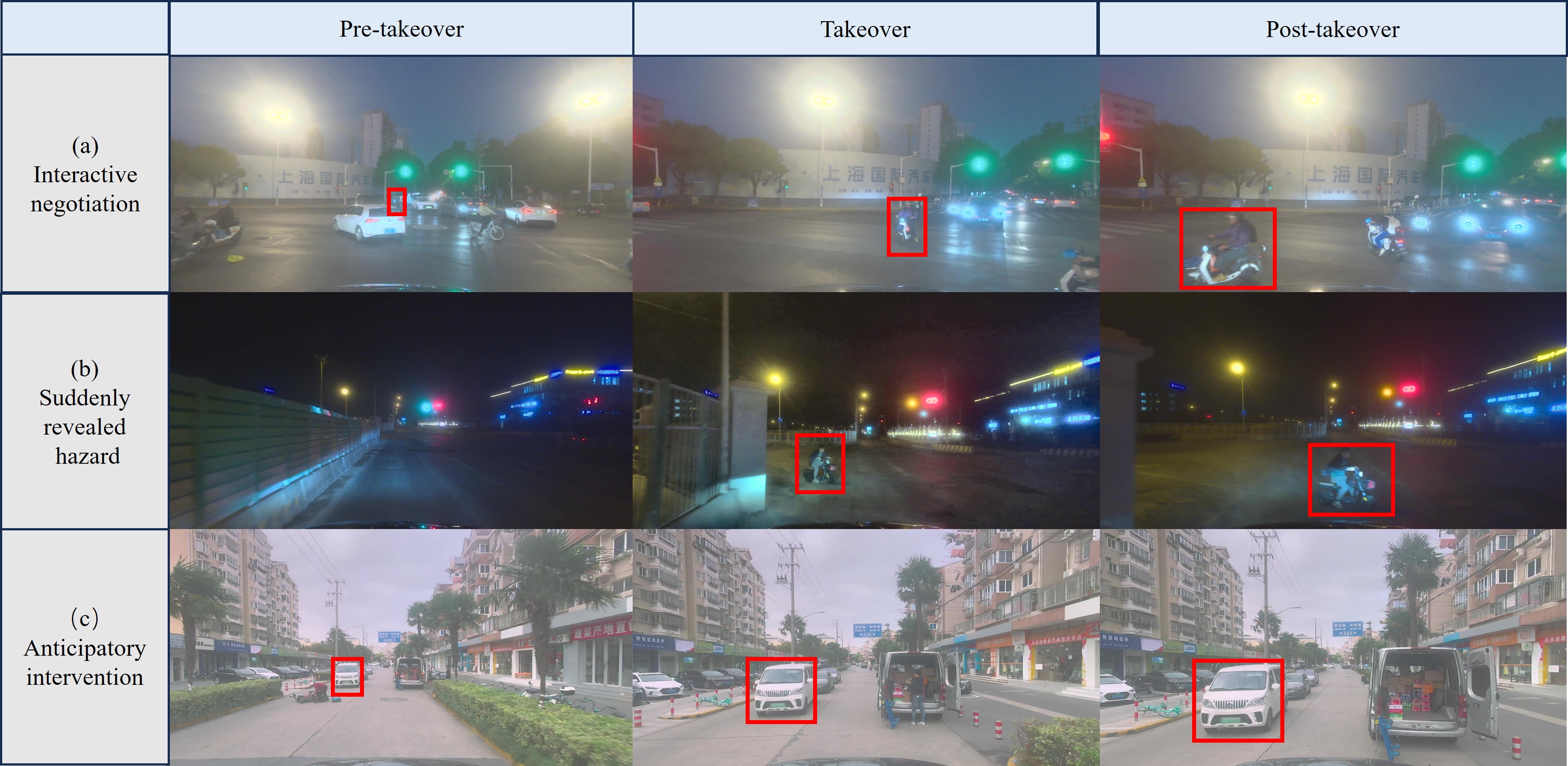}
    \caption{Representative interactive takeover events in InterScenes.
    The examples include (a) interaction negotiation, (b) a suddenly
    revealed hazard, and (c) anticipatory intervention. Each case shows
    selected observations before, at, and after takeover. Red boxes indicate
    the primary interacting agent.}
    \label{fig:takeover_cases}
\end{figure*}

\begin{table}[t]
\centering
\caption{Physical criticality of ordinary interactions and interactive
takeovers in InterScenes.}
\label{tab:takeover_criticality}
\scriptsize
\setlength{\tabcolsep}{4.5pt}
\renewcommand{\arraystretch}{1.10}
\begin{tabular}{lcc}
\toprule
\textbf{Metric} &
\textbf{Ordinary} &
\textbf{Takeover} \\
\midrule
Median minTTC (s)                & 2.986 & 0.965 \\
Median minimum OBB clearance (m) & 2.401 & 1.366 \\
\bottomrule
\end{tabular}
\end{table}

\subsection{Interactive Takeover Analysis}

InterScenes covers diverse road topologies and interaction
patterns, including car-following and head-on encounters
for which a unique shared conflict point may not be well
defined. For the within-dataset comparison, we therefore
use minimum two-dimensional time-to-collision (minTTC)
and minimum oriented bounding box (minOBB) clearance between
the ego vehicle and the primary interacting agent.
These metrics complement the conflict-point-based analysis
by characterizing predicted collision time margins and
spatial separation. As shown in Fig.~\ref{fig:ordinary_vs_takeover_ttc_obb} and
Table~\ref{tab:takeover_criticality}, the median minTTC
over valid episodes is 2.986~s for ordinary interactions
and 0.965~s for interactive takeovers. The corresponding
median minOBB clearances are 2.401~m and 1.366~m,
respectively. These results indicate that interactive
takeover episodes exhibit smaller spatial separation and,
among episodes with valid minTTC, shorter predicted
collision time margins.

Interestingly, however, takeover does not necessarily imply
high physical risk. Two characteristic intervention patterns
are observed. First, \emph{reactive takeovers} occur after challenging
interaction conditions have developed into small safety margins,
potentially exposing limitations in the ADS response to the interaction.
As shown in Fig.~\ref{fig:takeover_cases}(a)--(b), such cases include
close interaction negotiation and sudden interactions emerging from
occlusion. Second, \emph{anticipatory takeovers} occur before
trajectory-derived risk measures become critical.
In Fig.~\ref{fig:takeover_cases}(c), the safety driver
intervenes while physical margins remain relatively large.
The driver's interview account indicates that the intervention
was motivated by the anticipated development of the interaction.
These observations suggest that takeover events capture
both observed physical criticality and anticipated risk
perceived by the driver, providing information complementary
to trajectory-derived measures.

\section{Benchmarks and Experiments}
\label{sec:benchmarks}
We evaluate InterScenes from three perspectives:
(i) whether the automatically generated semantic annotations are reliable
and benefit from trajectory enhancement,
(ii) whether InterScenes, as an interaction-centered multimodal dataset,
poses greater challenges for VLA-based trajectory planning, and
(iii) whether physically grounded interaction-evolution descriptions
improve the physical fidelity of semantic-guided scenario generation
and support proving-ground deployment.

\subsection{Semantic Annotation Validation}
\label{subsec:semantic_validation}

\begin{table}[t]
\centering
\caption{Validation and ablation results for trajectory-enhanced semantic annotation (\%).}
\label{tab:semantic_ablation}
\small
\setlength{\tabcolsep}{1.8pt}
\renewcommand{\arraystretch}{1.12}
\begin{tabular}{@{}lcccc@{}}
\toprule
Method &
\shortstack{Road\\Topology $\uparrow$} &
\shortstack{Primary\\Agent $\uparrow$} &
\shortstack{Interaction\\Evolution $\uparrow$} &
\shortstack{Overall\\Acceptance $\uparrow$} \\
\midrule
w/o Traj. Enh. &
89.5 & 81.4 & 73.4 & 68.8 \\

Full &
\textbf{94.9} &
\textbf{91.1} &
\textbf{88.6} &
\textbf{85.2} \\
\bottomrule
\end{tabular}
\end{table}

We evaluate the full pipeline and the trajectory-enhancement
ablation on the same set of InterScenes episodes. Annotation
reliability is assessed against the corresponding visual
observations and aligned trajectories. Two human annotators
independently review each annotation with respect to road
topology, the primary interacting agent, and interaction
evolution, with disagreements resolved through discussion.
Interaction evolution is considered correct when the described
relative motion, yielding or passing order, and temporal
development are consistent with the observed interaction.
Minor linguistic variations that preserve the interaction
semantics are not treated as errors. An episode is accepted
overall only when all three criteria are satisfied.
The ``w/o Traj. Enh.'' setting removes all trajectory inputs
and trajectory-derived textual priors, leaving the VLM to
generate annotations from visual observations alone.

As shown in Table~\ref{tab:semantic_ablation}, the Full pipeline achieves
accuracies ranging from 88.6\% to 94.9\% across the three verification
criteria, together with an overall acceptance rate of 85.2\%. These results
indicate that the generated annotations are generally consistent with the
underlying visual observations and trajectories. Removing trajectory
enhancement reduces the overall acceptance rate to 68.8\% and decreases
the interaction-evolution accuracy from 88.6\% to 73.4\%. This result
further confirms the contribution of trajectory-derived information to
reliable interaction-evolution annotation. Annotations identified as
inconsistent are manually corrected before use in subsequent experiments.

\begin{table}[!t]
 \centering
\caption{Cross-dataset trajectory-planning comparison using OpenEMMA
with Qwen3-VL-8B.}
 \label{tab:vla_cross_dataset}
 \footnotesize
 \setlength{\tabcolsep}{2.2pt}
 
 \begin{tabular*}{\columnwidth}
 {@{\extracolsep{\fill}}llcccccc@{}}
 \hline
 
 Dataset &
 Method &
 \multicolumn{3}{c}{L2 (m)} &
 Avg. &
 Fail. &
 Ov. \\
 
 \cline{3-5}
 
 & & 1~s & 2~s & 3~s & & & \\
 
 \hline
 
 \multirow{2}{*}{nuScenes}
 & CV
 & 0.303
 & 0.676
 & 1.187
 & 0.722
 & 0.50\%
 & 0.582 \\
 
 & OpenEMMA
 & 0.302
 & 0.612
 & 1.044
 & 0.653
 & 1.78\%
 & 0.683 \\
 
 \hline
 
 \multirow{2}{*}{InterScenes}
 & CV
 & 0.310
 & 0.754
 & 1.368
 & 0.811
 & 1.56\%
 & 0.634 \\
 
 & OpenEMMA
 & 0.503
 & 0.990
 & 1.611
 & 1.035
 & 6.22\%
 & 0.983 \\
 
 \hline
 \end{tabular*}
 
 \end{table}

\subsection{VLA-Based Trajectory Planning Evaluation}
\label{subsec:vla_eval}

\subsubsection{Protocol and Metrics}

We adopt OpenEMMA~\cite{openemma} as the primary VLA planner and
constant velocity (CV) as a kinematic baseline. OpenEMMA predicts
future speed--curvature actions from the current front-view image and
historical ego states, while CV extrapolates ego motion with constant
speed and heading. We randomly select 100 cases from all InterScenes
episodes to form the evaluation subset and evaluate them at 2~Hz using
10 historical and 10 future samples with a stride of one, while the same protocol is applied to the 150-scene nuScenes validation split. Following OpenEMMA,
we report L2 errors at 1, 2, and 3~s, Avg L2, and Failure Rate. In our
implementation, the L2 metric at each horizon is the mean displacement
error over all predicted endpoints up to that horizon, and Avg L2 is the
mean of the three horizon-wise values. Failure Rate is the percentage of
predictions with a displacement error of at least 10~m at any point within
the first 3~s. We additionally report Overlap L2, which measures replanning
consistency by comparing consecutive predictions at shared future
timestamps, with lower values indicating more temporally consistent
planning.

\subsubsection{Cross-Dataset Comparison}

Under the unified evaluation protocol, we compare trajectory-planning
performance on the InterScenes evaluation subset and
nuScenes~\cite{nuScenes}, using Qwen3-VL-8B as the OpenEMMA backbone.
As shown in Table~\ref{tab:vla_cross_dataset}, both CV and OpenEMMA
exhibit higher trajectory errors on InterScenes. The Avg L2 of OpenEMMA
increases from 0.653~m on nuScenes to 1.035~m on InterScenes,
corresponding to a 58.5\% increase, while that of CV increases from
0.722~m to 0.811~m, corresponding to a 12.3\% increase. The Failure
Rate and Overlap L2 of OpenEMMA also increase from 1.78\% to 6.22\%
and from 0.683~m to 0.983~m, respectively. Notably, OpenEMMA achieves
a lower Avg L2 than CV on nuScenes, whereas this advantage disappears
on InterScenes. Overall, these results indicate that InterScenes poses
greater challenges for VLA-based trajectory planning. This observation
is also consistent with the larger proportion of interactions with
limited temporal safety margins discussed in
Sec.~\ref{subsec:interaction_criticality}.

\subsubsection{Effect of Multimodal Backbones}

\begin{table*}[!t]
\centering
\vspace*{3mm}
\caption{Planning performance and kinematic diagnostics of different
OpenEMMA backbones on the InterScenes evaluation subset.}
\label{tab:vla_backbone_100case}

\scriptsize
\renewcommand{\arraystretch}{1.05}

\begin{tabular*}{\textwidth}{@{\extracolsep{\fill}}lccccccccccc@{}}
\toprule

\multirow{3}{*}{Backbone} &
\multicolumn{6}{c}{Overall Planning Performance} &
\multicolumn{3}{c}{Interaction-wise Avg L2 (m) $\downarrow$} &
\multicolumn{2}{c}{Kinematic Diagnostics} \\

\cmidrule(lr){2-7}
\cmidrule(lr){8-10}
\cmidrule(lr){11-12}

&
\multicolumn{3}{c}{L2 (m) $\downarrow$} &
\multirow{2}{*}{Avg. $\downarrow$} &
\multirow{2}{*}{Fail. $\downarrow$} &
\multirow{2}{*}{Ov. $\downarrow$} &
\multirow{2}{*}{Car-Follow.} &
\multirow{2}{*}{Merging} &
\multirow{2}{*}{Crossing} &
\multirow{2}{*}{Speed MAE (m/s) $\downarrow$} &
\multirow{2}{*}{Heading MAE (rad) $\downarrow$} \\

\cmidrule(lr){2-4}

&
1~s &
2~s &
3~s &
&
&
&
&
&
&
&
\\

\midrule

Qwen3-VL-8B &
0.503 &
0.990 &
1.611 &
1.035 &
6.22\% &
0.983 &
0.577 &
0.790 &
1.885 &
1.463 &
0.123 \\

Qwen3-VL-32B &
0.563 &
1.095 &
1.766 &
1.141 &
7.33\% &
1.120 &
0.692 &
0.811 &
2.084 &
1.509 &
0.117 \\

Qwen3.8-27B &
\textbf{0.265} &
\textbf{0.628} &
\textbf{1.153} &
\textbf{0.682} &
\textbf{2.78\%} &
\textbf{0.700} &
\textbf{0.337} &
\textbf{0.653} &
\textbf{1.081} &
\textbf{1.253} &
\textbf{0.107} \\

\bottomrule
\end{tabular*}
\end{table*}

\begin{figure}[t]
    \centering
    \vspace*{3mm}\includegraphics[width=\columnwidth]{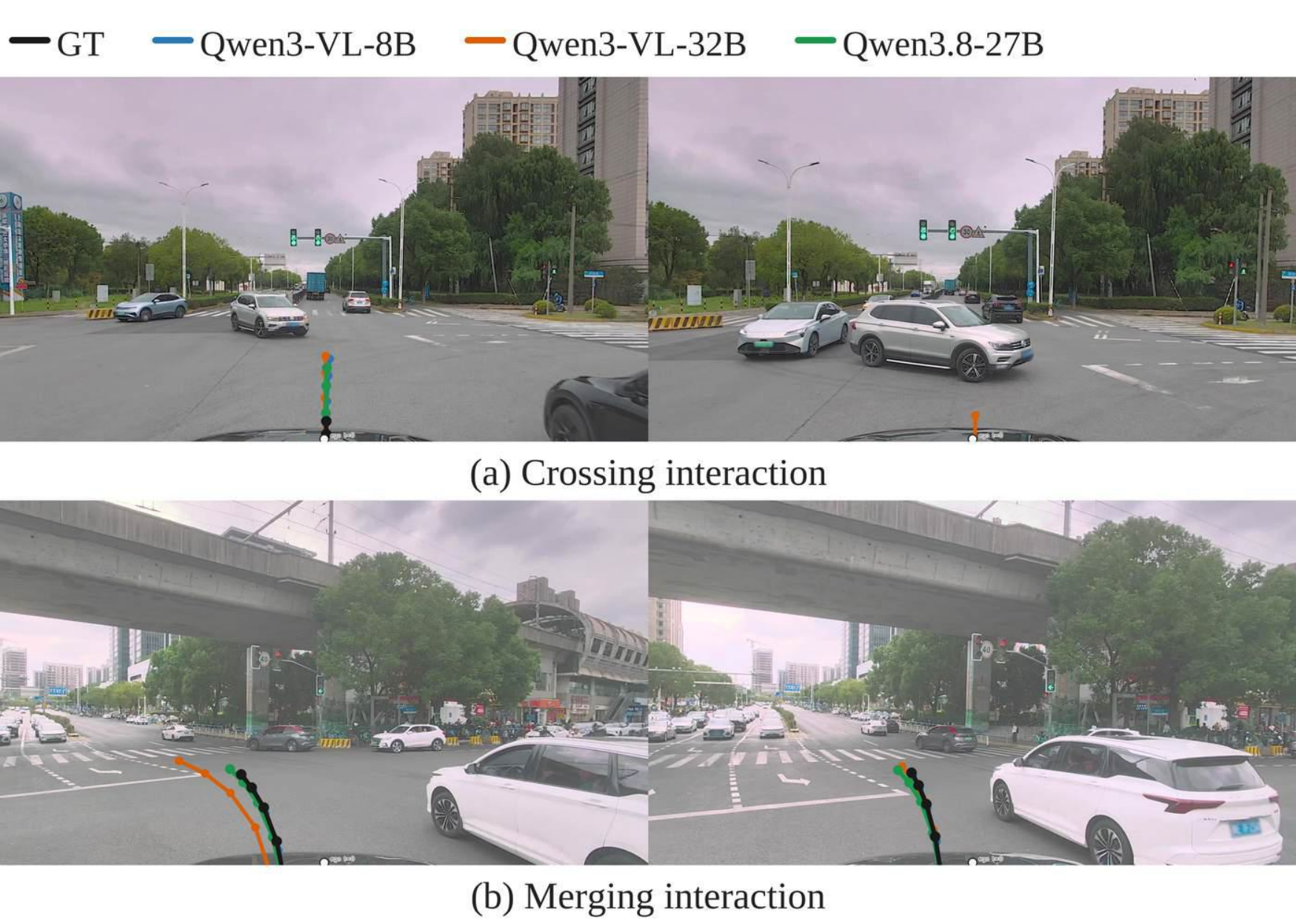}
    \caption{Qualitative comparison of backbone-dependent VLA trajectory
    planning on representative crossing and merging interactions. Each row
    shows two interaction stages from the same case. Black, blue, orange,
    and green denote GT, Qwen3-VL-8B, Qwen3-VL-32B, and Qwen3.8-27B,
    respectively.}
    \label{fig:vla_planning_qualitative}
\end{figure}

We compare three multimodal backbones under the same OpenEMMA planning
interface on the common InterScenes evaluation subset. As shown in
Table~\ref{tab:vla_backbone_100case}, Qwen3.8-27B achieves the best
overall performance, reducing Avg L2 from 1.035\,m with Qwen3-VL-8B
to 0.682\,m, while also obtaining the lowest Failure Rate and Overlap
L2. Increasing Qwen3-VL from 8B to 32B does not yield a consistent
improvement, with Avg L2 increasing to 1.141\,m. The performance gap is
also interaction dependent, with crossing showing the largest Avg L2
for all three backbones, while Qwen3.8-27B reduces the crossing error
from 1.885\,m to 1.081\,m. Head-on interactions are omitted from the
interaction-wise comparison because only a small number of such cases
are included in the randomly sampled evaluation subset, which is
insufficient for a representative comparison. Kinematic diagnostics
further show that Qwen3.8-27B achieves the lowest Speed MAE and Heading
MAE, whereas Qwen3-VL-32B obtains a lower Heading MAE than Qwen3-VL-8B
despite larger trajectory errors, indicating that improvements in
individual kinematic components do not necessarily translate into lower
overall trajectory error. Fig.~\ref{fig:vla_planning_qualitative}
provides qualitative examples in crossing and merging interactions,
illustrating backbone-dependent trajectory deviations from the ground
truth across different interaction stages.

\subsection{Semantic-Guided Scenario Generation and Deployment}
\label{subsec:scenario_generation}

We further evaluate the utility of InterScenes annotations from two
complementary perspectives: whether they can guide the generation of
executable simulation scenarios that preserve the physical
characteristics of the source interactions, and whether they can
support scenario deployment in a physical environment. For each source
interaction, the \emph{Base Description} provides basic scene and
behavior semantics, similar to a functional-level scenario
description~\cite{chat2scenic}, while the \emph{Interaction-Evolution
(IE) Description} provides a physically grounded description of
interaction evolution. We compare the Base and IE descriptions through semantic-guided scenario
generation in CARLA, and further evaluate IE-based proving-ground deployment.

\begin{figure}[!t]
    \centering
    \vspace*{1mm}\includegraphics[width=\columnwidth]{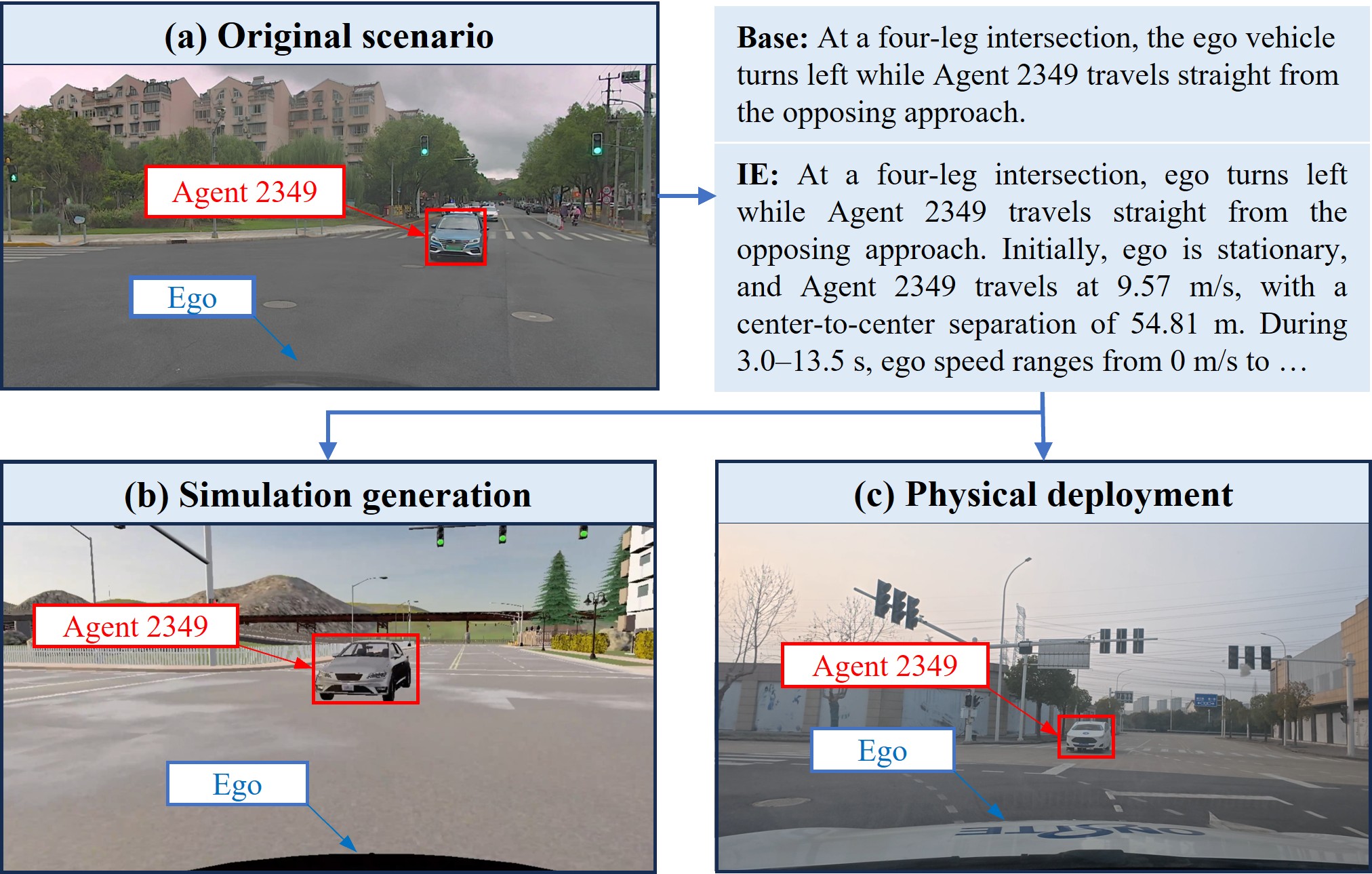}
 \caption{Semantic-guided scenario generation in CARLA and proving-ground
deployment from the same InterScenes interaction.}
    \label{fig:scenario_deployment}
\end{figure}

\begin{table}[!t]
\centering
\caption{Scenario-generation performance using Base and IE descriptions.}
\label{tab:scenario_generation}
\footnotesize
\setlength{\tabcolsep}{3.0pt}
\renewcommand{\arraystretch}{1.05}

\begin{tabular}{llccc}
\toprule
Generator & Description &
CSR (\%) $\uparrow$ &
SA (\%) $\uparrow$ &
PFS $\uparrow$ \\
\midrule

\multirow{2}{*}{ChatScene}
& Base & \textbf{25.7} & \textbf{22.2} & 65.5 \\
& IE   & 14.3 & 20.0 & \textbf{79.7} \\

\midrule

\multirow{2}{*}{Chat2Scenic}
& Base & \textbf{64.3} & 24.4 & 74.1 \\
& IE   & 54.3 & \textbf{34.2} & \textbf{82.2} \\

\bottomrule
\end{tabular}
\end{table}

\subsubsection{Scenario Generation in Simulation}

We randomly select 70 InterScenes cases for semantic-to-scenario
generation in CARLA using ChatScene~\cite{chatscene} and
Chat2Scenic~\cite{chat2scenic}. The same 70 cases are used across
all generators and description settings. For each source interaction,
the Base Description and IE Description are evaluated under the same
generation setting. We report compilation and execution success rate (CSR), semantic
alignment (SA), and physical fitting score (PFS). CSR measures the
percentage of source cases for which the generated scenario can be
successfully compiled and executed. SA measures the percentage of successfully executed scenarios
that are consistent with the source interaction in all four
criteria: road topology, ego behavior, primary-agent behavior,
and interaction relation. PFS is computed only for successfully
executed scenarios that satisfy SA and have valid trajectories.
It measures physical consistency with the source interaction
using the initial relative-position error $E_p$, ego speed
MAE $E_e$, and primary-agent speed MAE $E_a$:

\begin{equation}
\hat{E}_{k}
=
\min\left(
\frac{E_k}{\tau_k},
1
\right),
\quad
k \in \{p,e,a\},
\end{equation}
\begin{equation}
\mathrm{PFS}_{i}
=
100\left[
1-
\frac{
\hat{E}_{p}^{i}
+
\hat{E}_{e}^{i}
+
\hat{E}_{a}^{i}
}{3}
\right].
\end{equation}
Here, $\tau_k$ denotes a fixed normalization scale
determined as the 95th percentile of the corresponding
error distribution pooled across all generators and
description settings. The scales are computed once
and shared across all settings.

As illustrated in Fig.~\ref{fig:scenario_deployment}(a) and (b), the
IE Description guides CARLA scenario generation while preserving the
characteristic interaction process of the source case. As shown in Table~\ref{tab:scenario_generation}, adding IE information
reduces CSR from 25.7\% to 14.3\% for ChatScene and from 64.3\% to
54.3\% for Chat2Scenic, showing that the additional physical and
interaction-evolution constraints reduce the success rate of executable
scenario generation.
Meanwhile, PFS increases from 65.5 to 79.7 for ChatScene and from
74.1 to 82.2 for Chat2Scenic, demonstrating that interaction-evolution
descriptions improve the physical fidelity of semantic-guided scenario
generation. The effect on semantic alignment is generator dependent:
SA increases from 24.4\% to 34.2\% for Chat2Scenic but slightly
decreases from 22.2\% to 20.0\% for ChatScene. Overall, interaction-evolution descriptions improve the physical fidelity
of semantic-guided scenario generation, while resulting in lower CSR and
generator-dependent changes in semantic alignment.

\subsubsection{Scenario Deployment in Physical Environments}

We further evaluate whether the IE Description can support
scenario deployment in a physical testing environment.
This experiment focuses on the IE Description, which
provides physical and temporal constraints for deployment.
We deploy two representative InterScenes cases using
Video2Track~\cite{video2track}, with five repeated trials
for each case. The road topology specified in the IE
Description is matched to the proving-ground map, and
corresponding spatial anchors, including participant
starting positions and conflict points, are selected.
Conditioned on these anchors and interaction constraints,
a diffusion model generates candidate trajectories.
Stackelberg-game optimization then refines the trajectories
to align their minTTC with the target specified by the
IE Description, yielding executable trajectories for
closed-track testing.

As illustrated in Fig.~\ref{fig:scenario_deployment}(a) and (c),
the deployed scenarios preserve the source interaction
types and characteristic interaction processes.
Across all 10 physical trials, the relative minTTC error
with respect to the source interaction remains within 10\%.
These results provide feasibility evidence that IE
Descriptions can guide topology matching, spatial-anchor
selection, and trajectory generation for proving-ground
scenario deployment.

\section{Conclusion}

This paper presents InterScenes, a real-world multimodal
interaction-evolution dataset for VLA evaluation and ADS safety
validation in interaction-intensive urban traffic. InterScenes integrates
synchronized multimodal observations, ego and surrounding-agent
trajectories, physically grounded interaction-evolution descriptions,
and real-world interactive takeover events, representing traffic
interactions as physically grounded temporal processes. Dataset analysis
shows substantial coverage of temporally close interactions, with 11.6\% and 24.2\% of episodes having 
a $\Delta TTCP_{\min}$
below 1~s and 3~s, respectively. Interactive takeovers exhibit
smaller minOBB clearances and, among episodes with valid
minTTC, shorter predicted collision time margins than ordinary
interactions. Representative cases also show interventions while
physical margins remain relatively large, suggesting that takeover
events provide complementary information beyond trajectory-derived
criticality measures.
Cross-dataset VLA evaluation further shows that InterScenes poses greater
challenges for trajectory planning, while the backbone comparison reveals
clear performance differences among multimodal models. The
interaction-evolution annotations also improve the physical fidelity
of semantic-guided scenario generation in simulation environments and
provide physically grounded constraints for deployment in physical
testing environments.

Future work will expand InterScenes to more cities, regions, and road
environments, including rural roads, while evaluating more VLA models
and extending interaction-evolution semantics to larger-scale closed-loop
scenario generation and ADS safety validation.








\bibliographystyle{IEEEtran}
\bibliography{ref}

@InProceedings{WOMD,
    author    = {Ettinger, Scott and Cheng, Shuyang and Caine, Benjamin and Liu, Chenxi and Zhao, Hang and Pradhan, Sabeek and Chai, Yuning and Sapp, Ben and Qi, Charles R. and Zhou, Yin and Yang, Zoey and Chouard, Aur\'elien and Sun, Pei and Ngiam, Jiquan and Vasudevan, Vijay and McCauley, Alexander and Shlens, Jonathon and Anguelov, Dragomir},
    title     = {Large Scale Interactive Motion Forecasting for Autonomous Driving: The Waymo Open Motion Dataset},
    booktitle = {Proceedings of the IEEE/CVF International Conference on Computer Vision (ICCV)},
    year      = {2021},
    pages     = {9710-9719}
}

@misc{INTERACTION,
      title={INTERACTION Dataset: An INTERnational, Adversarial and Cooperative moTION Dataset in Interactive Driving Scenarios with Semantic Maps}, 
      author={Wei Zhan and Liting Sun and Di Wang and Haojie Shi and Aubrey Clausse and Maximilian Naumann and Julius Kummerle and Hendrik Konigshof and Christoph Stiller and Arnaud de La Fortelle and Masayoshi Tomizuka},
      year={2019},
      eprint={1910.03088},
      archivePrefix={arXiv},
      primaryClass={cs.RO},
      url={https://arxiv.org/abs/1910.03088}, 
}

@InProceedings{nuScenes,
author = {Caesar, Holger and Bankiti, Varun and Lang, Alex H. and Vora, Sourabh and Liong, Venice Erin and Xu, Qiang and Krishnan, Anush and Pan, Yu and Baldan, Giancarlo and Beijbom, Oscar},
title = {nuScenes: A Multimodal Dataset for Autonomous Driving},
booktitle = {Proceedings of the IEEE/CVF Conference on Computer Vision and Pattern Recognition (CVPR)},
month = {June},
year = {2020}
}

@article{interhub,
  title={A naturalistic trajectory dataset with dense interaction for autonomous driving},
  author={Jiang, Xiyan and Zhao, Xiaocong and Liu, Yiru and Li, Zirui and Hang, Peng and Xiong, Lu and Sun, Jian},
  journal={Scientific Data},
  volume={12},
  pages={1084},
  year={2025},
  doi={10.1038/s41597-025-05344-7}
}

@misc{liu2025autonomous,
  author       = {Liu, Henry and Cao, Zhong and Yan, Xintao and Feng, Shuo and Lu, Qiujing},
  title        = {Autonomous Vehicles: A Critical Review (2004--2024) and a Vision for the Future},
  year         = {2025},
  month        = may,
  day          = {30},
  publisher    = {TechRxiv},
  doi          = {10.36227/techrxiv.174857767.78237989/v1},
  url          = {https://www.techrxiv.org/doi/abs/10.36227/techrxiv.174857767.78237989/v1}
}

@ARTICLE{zhaoyongqisurvey,
  author={Zhao, Yongqi and Zhou, Ji and Bi, Dong and Mihalj, Tomislav and Hu, Jia and Eichberger, Arno},
  journal={IEEE Transactions on Intelligent Transportation Systems}, 
  title={A Survey on the Application of Large Language Models in Scenario-Based Testing of Automated Driving Systems}, 
  year={2026},
  volume={},
  number={},
  pages={1-23},
  doi={10.1109/TITS.2026.3651004}}

@misc{dsl2021,
  author = {{The Standing Committee of the National People's Congress of the People's Republic of China}},
  title  = {{Data Security Law of the People's Republic of China}},
  year   = {2021},
  note   = {Order No. 84, effective Sept. 1, 2021}
}

@software{dashcam_anonymizer,
  author = {Varun Gupta},
  month = {8},
  title = {{Dashcam Anonymizer}},
  url = {https://github.com/varungupta31/dashcam_anonymizer},
  version = {1.0.0},
  year = {2023}
}

@INPROCEEDINGS{openemma,
  author={Xing, Shuo and Qian, Chengyuan and Wang, Yuping and Hua, Hongyuan and Tian, Kexin and Zhou, Yang and Tu, Zhengzhong},
  booktitle={2025 IEEE/CVF Winter Conference on Applications of Computer Vision Workshops (WACVW)}, 
  title={OpenEMMA: Open-Source Multimodal Model for End-to-End Autonomous Driving}, 
  year={2025},
  volume={},
  number={},
  pages={911-919},
  doi={10.1109/WACVW65960.2025.00113}}

@misc{video2track,
      title={Video2Track: From Real-World Interaction Videos to Steerable Adversarial Closed-Track Testing for Automated Driving Systems}, 
      author={Mengjie Tian and Xinrui Zhang and Tianyu Li and Peizhi Zhang and Guirong Zhou and Haojie Feng and Junpeng Huang and Qixiang Zhang and Lu Xiong},
      year={2026},
      eprint={2608.11592},
      archivePrefix={arXiv},
      primaryClass={cs.RO},
      url={https://arxiv.org/abs/2608.11592}, 
}

@INPROCEEDINGS{chatscene,
  author={Zhang, Jiawei and Xu, Chejian and Li, Bo},
  booktitle={2024 IEEE/CVF Conference on Computer Vision and Pattern Recognition (CVPR)}, 
  title={ChatScene: Knowledge-Enabled Safety-Critical Scenario Generation for Autonomous Vehicles}, 
  year={2024},
  volume={},
  number={},
  pages={15459-15469},
  doi={10.1109/CVPR52733.2024.01464}}

@inproceedings{CCD,
author = {Bao, Wentao and Yu, Qi and Kong, Yu},
title = {Uncertainty-based Traffic Accident Anticipation with Spatio-Temporal Relational Learning},
year = {2020},
isbn = {9781450379885},
publisher = {Association for Computing Machinery},
address = {New York, NY, USA},
url = {https://doi.org/10.1145/3394171.3413827},
doi = {10.1145/3394171.3413827},
booktitle = {Proceedings of the 28th ACM International Conference on Multimedia},
pages = {2682–2690},
numpages = {9},
location = {Seattle, WA, USA},
series = {MM '20}
}

@InProceedings{trafficalign,
    author    = {Tu, Zhi and Niu, Liangkun and Zhang, Tianyi},
    title     = {TrafficAlign: Aligning Large Language Models for Traffic Scenario Generation},
    booktitle = {Proceedings of the IEEE/CVF Conference on Computer Vision and Pattern Recognition (CVPR)},
    month     = {June},
    year      = {2026},
    pages     = {39744-39754}
}

@INPROCEEDINGS{SIND,
  author={Xu, Yanchao and Shao, Wenbo and Li, Jun and Yang, Kai and Wang, Weida and Huang, Hua and Lv, Chen and Wang, Hong},
  booktitle={2022 IEEE 25th International Conference on Intelligent Transportation Systems (ITSC)}, 
  title={SIND: A Drone Dataset at Signalized Intersection in China}, 
  year={2022},
  volume={},
  number={},
  pages={2471-2478},
  doi={10.1109/ITSC55140.2022.9921959}}

@InProceedings{VLA4CoDrive,
    author    = {Boroujeni, Sayed Pedram Haeri and Razi, Abolfazl},
    title     = {VLA4CoDrive: Vision-Language-Action Dataset for Cooperative Autonomous Driving},
    booktitle = {Proceedings of the IEEE/CVF Winter Conference on Applications of Computer Vision (WACV) Workshops},
    month     = {March},
    year      = {2026},
    pages     = {1789-1799}
}

@techreport{nhtsa_sgo,
  author      = {{National Highway Traffic Safety Administration}},
  title       = {Summary Report: Standing General Order on Crash Reporting for Automated Driving Systems},
  institution = {U.S. Department of Transportation, National Highway Traffic Safety Administration},
  number      = {DOT HS 813 324},
  month       = {June},
  year        = {2022}
}

@misc{chat2scenic,
      title={Chat2Scenic: An Iterative RAG-Based Framework for Scenario Generation in Autonomous Driving}, 
      author={Yuan Gao and Wenting Miao and Mattia Piccinini and Haoyu Wang and Qunying Song and Johannes Betz},
      year={2026},
      eprint={2607.14387},
      archivePrefix={arXiv},
      primaryClass={cs.AI},
      url={https://arxiv.org/abs/2607.14387}, 
}

@INPROCEEDINGS{WOD,
  author={Sun, Pei and Kretzschmar, Henrik and Dotiwalla, Xerxes and Chouard, Aurélien and Patnaik, Vijaysai and Tsui, Paul and Guo, James and Zhou, Yin and Chai, Yuning and Caine, Benjamin and Vasudevan, Vijay and Han, Wei and Ngiam, Jiquan and Zhao, Hang and Timofeev, Aleksei and Ettinger, Scott and Krivokon, Maxim and Gao, Amy and Joshi, Aditya and Zhang, Yu and Shlens, Jonathon and Chen, Zhifeng and Anguelov, Dragomir},
  booktitle={2020 IEEE/CVF Conference on Computer Vision and Pattern Recognition (CVPR)}, 
  title={Scalability in Perception for Autonomous Driving: Waymo Open Dataset}, 
  year={2020},
  volume={},
  number={},
  pages={2443-2451},
  doi={10.1109/CVPR42600.2020.00252}}

@ARTICLE{ARTEMIS-3,
  author={Feng, Renju and Xi, Ning and Chu, Duanfeng and Wang, Rukang and Deng, Zejian and Wang, Anzheng and Lu, Liping and Wang, Jinxiang and Huang, Yanjun},
  journal={IEEE Robotics and Automation Letters}, 
  title={ARTEMIS: Autoregressive End-to-End Trajectory Planning With Mixture of Experts for Autonomous Driving}, 
  year={2026},
  volume={11},
  number={1},
  pages={226-233},
  doi={10.1109/LRA.2025.3632616}}

@misc{yu2026ssquaredvladecouplingsemanticspatial,
      title={S-squared-VLA: Decoupling Semantic and Spatial Streams in Vision-Language-Action Models for Autonomous Driving}, 
      author={Jianguo Yu and Rukang Wang and Duanfeng Chu and Chen Wang and Renju Feng and Liping Lu},
      year={2026},
      eprint={2607.13926},
      archivePrefix={arXiv},
      primaryClass={cs.RO},
      url={https://arxiv.org/abs/2607.13926}, 
}

@misc{gao2026nuriskvisualquestionanswering,
      title={NuRisk: A Visual Question Answering Dataset for Agent-Level Risk Assessment in Autonomous Driving}, 
      author={Yuan Gao and Mattia Piccinini and Roberto Brusnicki and Yuchen Zhang and Johannes Betz},
      year={2026},
      eprint={2509.25944},
      archivePrefix={arXiv},
      primaryClass={cs.AI},
      url={https://arxiv.org/abs/2509.25944}, 
}

@InProceedings{wod_e2e,
    author    = {Xu, Runsheng and Lin, Hubert and Jeon, Wonseok and Feng, Hao and Zou, Yuliang and Sun, Liting and Gorman, John and Tolstaya, Kate and Tang, Sarah and White, Brandyn and Sapp, Ben and Tan, Mingxing and Hwang, Jyh-Jing and Anguelov, Dragomir},
    title     = {WOD-E2E: Waymo Open Dataset for End-to-End Driving in Challenging Long-tail Scenarios},
    booktitle = {Proceedings of the IEEE/CVF Conference on Computer Vision and Pattern Recognition (CVPR)},
    month     = {June},
    year      = {2026},
    pages     = {3709-3718}
}

@article{XU2026116002,
title = {DuSA: Dual-loop self-learning framework for autonomous driving with LLM-guided reinforcement learning},
journal = {Knowledge-Based Systems},
volume = {343},
pages = {116002},
year = {2026},
issn = {0950-7051},
doi = {https://doi.org/10.1016/j.knosys.2026.116002},
url = {https://www.sciencedirect.com/science/article/pii/S0950705126007288},
author = {Jinchang Xu and Sunan Zhang and Chen Sun and Jiawei Wang and Yang Xiao and Bingbing Li and Yulong Wang and Guodong Yin and Weichao Zhuang}
}

@INPROCEEDINGS{Drivellm,
  author={Chen, Long and Sinavski, Oleg and H{\"u}nermann, Jan and
          Karnsund, Alice and Willmott, Andrew James and Birch, Danny and
          Maund, Daniel and Shotton, Jamie},
  booktitle={2024 IEEE International Conference on Robotics and Automation (ICRA)}, 
  title={Driving with LLMs: Fusing Object-Level Vector Modality for Explainable Autonomous Driving}, 
  year={2024},
  volume={},
  number={},
  pages={14093-14100},
  doi={10.1109/ICRA57147.2024.10611018}}

@INPROCEEDINGS{Driveanywhere,
  author={Wang, Tsun-Hsuan and Maalouf, Alaa and Xiao, Wei and Ban, Yutong and Amini, Alexander and Rosman, Guy and Karaman, Sertac and Rus, Daniela},
  booktitle={2024 IEEE International Conference on Robotics and Automation (ICRA)}, 
  title={Drive Anywhere: Generalizable End-to-end Autonomous Driving with Multi-modal Foundation Models}, 
  year={2024},
  volume={},
  number={},
  pages={6687-6694},
  doi={10.1109/ICRA57147.2024.10611590}}

\end{document}